%% file: root.tex
\pdfoutput=1
\documentclass[letterpaper, 10 pt, conference]{ieeeconf}  

\IEEEoverridecommandlockouts                              

\usepackage[T1]{fontenc}

\usepackage{hyperref}
\usepackage[binary-units]{siunitx}
\usepackage{svg}
\usepackage{graphicx}
\graphicspath{{images/}}
\usepackage{amsmath}
\usepackage{makecell}
\usepackage{tabularx}
\usepackage{color,colortbl}
\usepackage{amssymb}
\usepackage[utf8]{inputenc}

\usepackage{lscape}
\usepackage{rotating,siunitx}
\usepackage{pdflscape}
\usepackage{rotating}
\usepackage{tabu}
\usepackage{array,url,kantlipsum}

\usepackage{makecell}

\usepackage{multicol,lipsum}
\usepackage{multirow}
\usepackage{newtxmath}
\usepackage{booktabs,ragged2e}
\usepackage[flushleft]{threeparttable}

\usepackage{textcomp}
\usepackage{tablefootnote}
\usepackage{hhline}
\usepackage{colortbl}
\usepackage{longtable}

\usepackage{scalerel}
\usepackage{tikz}
\usetikzlibrary{svg.path,shapes.geometric,arrows,arrows.meta,decorations.pathreplacing,angles,quotes,math}

\usepackage{pgfplots} 
\usepackage{pgfplotstable}
\usepgfplotslibrary{colormaps}
\usepackage{pgfgantt}
\usepgfplotslibrary{fillbetween}

\usepackage{bm}

\usepackage{etoolbox}
\patchcmd{\thebibliography}{\chapter*}{\par\let\clearpage\relax\chapter*}{\typeout{success}}{\typeout{failure}}
\renewcommand{\clearpage}{}

\DeclareMathOperator*{\argmin}{arg\,min}

\title{\LARGE \bf Model-Based Disturbance Estimation for a Fiber-Reinforced Soft Manipulator using Orientation Sensing}

\author{Barnabas Gavin Cangan$^{1*}$, Stefan Escaida Navarro$^{2*}$, Bai Yang$^{3}$, Yu Zhang$^{1}$, Christian Duriez$^{2}$,\\ Robert K. Katzschmann$^{1}$
\thanks{* denotes equal contribution}
\thanks{$^{1}$Soft Robotics Lab, ETH Zurich, Switzerland
        {\texttt{\small\{bcangan, zhangyu, rkk\}@ethz.ch }}}%
\thanks{$^{2}$INRIA, Lille, France
        {\texttt{\small\{stefan.escaida-navarro, christian.duriez\}@inria.fr }}}%
\thanks{$^{3}$TU Berlin, Germany
        {\texttt{\small baaiy0610@mail.tu-berlin.de }}}%
}

\begin{document}

\maketitle
\thispagestyle{empty}
\pagestyle{empty}


\begin{abstract}
\input{content/01-abstract}
\end{abstract}

\section{Introduction}
\input{content/02-introduction}

\section{Related Work}
\input{content/03-related-work}

\section{Numerical Modeling Techniques}
\input{content/05-modeling}

\section{Experiments}
\input{content/06-experiments}

\input{content/07-results}


\section{Conclusions and Future Work}
\input{content/09-conclusion}










\bibliographystyle{IEEEtran}
\include{Bibliography}








\end{document}

%% file: content/01-abstract.tex
%
%
%
%
A soft robotic arm should ideally be working efficiently, robustly, and safely in human-centered environments to provide true assistance in real world situations. For this goal, soft robots need to be be able to estimate their state and external interactions based on (proprioceptive) sensors. Estimating disturbances allows a soft robot to perform desirable force control. Even in the case of rigid manipulators, force estimation at the end-effector is seen as a non-trivial problem. And indeed, other current approaches to address this challenge have shortcomings that prevent their general application. They are often based on simplified soft dynamic models, such as the ones relying on a piece-wise constant curvature approximation or matched rigid-body models that do not represent enough details of the problem. Thus, the applications needed for complex human-robot interaction can not be built. 
%
%
%
%
%
%
Finite element methods (FEM) allow for predictions of soft robot dynamics in a more generic fashion. Here, using the soft robot modeling capabilities of the framework SOFA, we build an detailed FEM model of a multi-segment soft continuum robotic arm composed of compliant deformable materials, and fiber-reinforced pressurized actuation chambers. In addition, a model for sensors that provide orientation output is presented. This model is used to establish a state observer for the manipulator. The sensor model is adequate for representing the output of flexible bend-sensors as well as orientations provided by IMUs or coming from tracking systems, all of which are popular choices in soft robotics.
Model parameters were calibrated to match imperfections of the manual fabrication process using physical experiments. We then solve a quadratic programming inverse dynamics problem to compute the components of external force that explain the pose error. Our experiments show an average force estimation error of around 1.2\%.
%
%
As the methods proposed are generic, these results are encouraging for the task of building soft robots exhibiting complex, reactive, sensor-based behavior that can be deployed in human-centered environments.

%% file: content/02-introduction.tex
\subsection{Motivation}
\label{subsec:Motivation}
Robotic manipulators allow tasks to be automated with minimal task-specific hardware development. However, robots must become more skillful, adaptive, and safe to be fully generalizable to automate tasks in unstructured everyday environments.
Soft robotics takes inspiration from nature to create safer machines that can perform tasks which were previously difficult or impossible for conventional rigid-bodied robots~\cite{rus_design_2015,hawkes_hard_2021}. The body of a soft robot is inherently compliant, this is particularly useful for interactions with the environment~\cite{della_santina_dynamic_2018,homberg_haptic_2015,coad_vine_2020,n_kuppuswamy_soft-bubble_2021,lessing_soft_2018}. For example, soft robotic arms have been applied to medical procedures such as lung biopsies by adapting their shape to explore complex internal structures~\cite{graetzel_robotic_2019,rogers_methods_2018}. Their soft bodies could eventually enable safer, and thereby closer, human-robot physical interactions~\cite{nguyen_fabric_2019}.

\begin{figure}[!t]
    \centering
    \includegraphics[width = 0.95\linewidth,viewport=0 0 3333 2783]{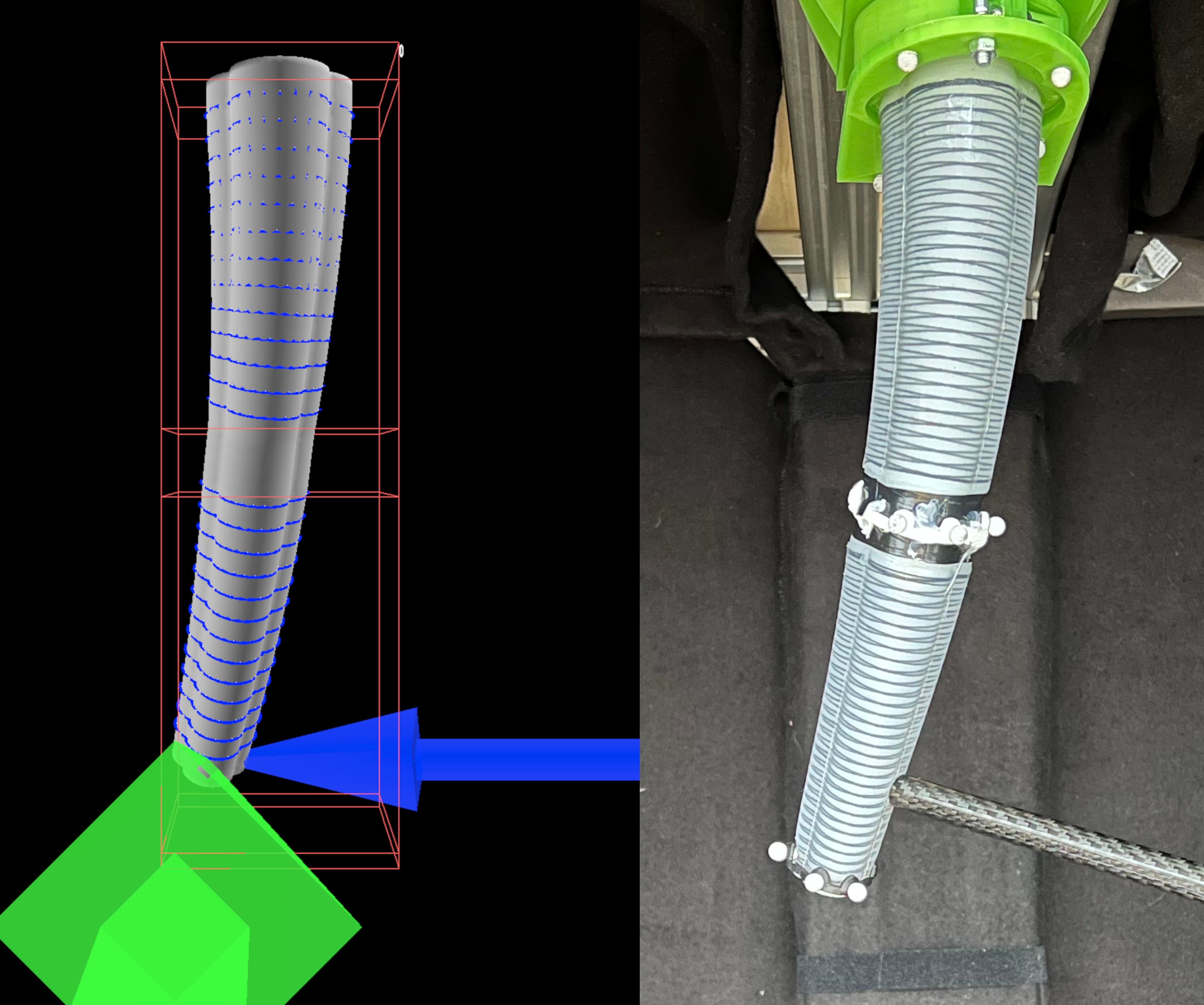}
    \caption{In this paper, we present the FEM-based modeling of a soft manipulator arm actuated by pressurized air. Sensing and actuation are coupled in the simulation to estimate disturbances on the arm. This is a building stone to achieve the next level of applications, such as the ones requiring kinesthetic control, which are not yet mature in soft robotics. The figure on the left shows the arm in simulation with components of estimated external forces acting at the tip. Green arrow pointing inward is the force component along the X-axis and blue arrow points represents force component along the Y-axis}
    \label{fig:overall-system}
\end{figure}

\subsection{Challenge}

In order to equip a soft robot with the means of estimating it's own state, including external forces, a complex interplay between different building blocks is necessary. Actuation, perception and modeling need to come together smoothly. The need for modeling approaches that are compatible with typical sensing and control cycles, have led to the use of simplified soft dynamic models, such as the ones relying on a piece-wise constant curvature approximation or matched rigid-body models. However, the limited modeling ability prevents building more complex soft robotic applications. Furthermore, data-driven methods are also a popular approach, but likewise are affected by problems that limit their applicability, such as data scarcity and lack of generality.



\subsection{Approach}
\label{subsec:Approach}
To address this problem, we propose a modeling approach that uses the FEM and is can thus provide a generic solution.

This paper builds upon previous work within the research group at ETH Zurich. In 
\cite{toshimitsu2021sopra}, Toshimitsu et al.\ introduced the \emph{Soft continuum Proprioceptive Arm} (SoPrA), which we use here. The design emphasizes proprioception, i.\,e.\ the ability of the arm to observe its state without external sensing, which is realized with capacitive orientation/bending sensors by BendLabs.\footnote{\url{https://www.bendlabs.com/products/2-axis-soft-flex-sensor}} In \cite{toshimitsu2021sopra}, the fabrication, sensorization, kinematics and an approach for dynamics modeling for SoPrA are presented. 

In addition, this work follows up on the results shown in \cite{navarro2020model,navarro2021bio} by the contributors from INRIA Lille, where an approaches for model-based sensing using the FEM in the framework SOFA was shown.\footnote{\url{https://www.sofa-framework.org/}} These approaches allow the seamless integration of actuation efforts and state estimation by the sensor systems. Thus, in this work, the mechanics SoPrA are modeled with the FEM. This addresses the actuation by pressurized chambers, the fiber reinforcement of these chambers as well as the use of an inverse model to estimate external disturbances~\cite{coevoet2017software}. To this end, the framework of inverse problem solving is extended with a generic sensor model for sensors that can output relative orientations between two arm segments (bending sensors, IMU, fiduciary marker tracking). In total, we show that the proposed approach allows us to address challenges, such as inverse dynamics that can be used in a teach-in mode as well as for force estimation.

popular sensing modality

We validate our approach in physical experiments on a soft robotic arm with multiple fiber-reinforced actuation chambers per soft robotic arm segment.

\subsection{Results summary}
Specifically, this work contributes:
\begin{enumerate}
    \item An FEM-based model for soft robotic manipulators that includes fiber reinforcement
    \item An inverse-dynamics-based calibration process to determine model parameters
    \item Validation of the calibration process
    \item A teach mode to demonstrate the effectiveness of the calibration approach
    \item Disturbance estimation at the tip with chambers not pressurized
    \item Disturbance estimation with the arm chambers pressurized.
\end{enumerate}


%% file: content/03-related-work.tex
In \cite{polygerinos2015modeling}, Polygerinos et al.\ derive an in depth model of fiber-reinforced soft pressurized bending actuators. However, this model cannot easily be used to capture the behavior of a robot design like SoPrA, featuring six such chambers, nor easily be extended for tasks such as inverse kinematics or disturbance estimation. SoPrA is a multi-segment arm with fiber reinforcements for dynamic object manipulation~\cite{toshimitsu2021sopra}. 

In \cite{tapia2020makesense}, Tapia et al.\ investigate the sensorization of soft robots using an approach, which allows to equip the robots with an optimal amount of stretch sensors needed to reconstruct its state. The robot's mechanics are modeled using the FEM and a model for the resistive type sensors is established, allowing to relate deformations to sensor values and vice-versa. This allows to find the deformations due to actuation efforts and external forces. However, this approach is lacking the possibility to localize the external contacts and thus the authors ``[...]  estimate contact forces of unknown magnitude and direction on the complete surface of the gripper.'' Based on the the previous work shown in \cite{navarro2020model}, we can state that the approach presented here is more generic and can integrate knowledge about the contact location, when acquired by a sensor skin, into the model.

In recent robotics literature, we can find approaches for orientation-based proprioceptive sensing. Thus, recent contributions show the integration of this type of sensing (IMUs) in deformable circuits, also featuring other sensing modalities \cite{yirmibesoglu2016hybrid,hellebrekers2018liquid}. In~\cite{hellebrekers2018liquid}, a single IMU is integrated into the skin covering a gripper. In~\cite{yirmibesoglu2016hybrid}, the proposed sensor, featuring one IMU at each end of a strain sensing element, is tested for instrumenting a fiber reinforced pneumatic actuator as well as for instrumenting adjacent links of a rigid robot. In terms of use of orientation information for state estimation, the work by Hughes et al.~\cite{hughes2020sensing} is closest to ours. The authors employ the orientation values provided by an IMU to estimate the pose of the tip of a single-segment continuum robot. The IMU allows to estimate the kinematic parameters of this segment using the Piecewise Constant-Curvature (PCC) modeling approach. While the authors show the application in closed-loop control, here, we show how this information can be used for inverse kinematics as well as disturbance estimation.

Data driven methods are also popular for soft robot state estimation. Capacitive sensing has been studied by Scimeca et al.~\cite{scimeca2019model}. The authors propose to use a capacitive tactile array at the base of a soft finger segment, sensing pressure distribution in order to estimate the pose of the tip using a feed-forward neural network trained using visual tracking data. However, this approach lacks methodology to estimate external forces to estimate the resulting deformations.

In \cite{della2020data}, Della Santina et al.\ demonstrated detecting external disturbances on a serial soft robotic manipulator similar to SoPrA. They designed non-linear observer based on the PCC approximation that would exactly compute external forces as long as the PCC assumption holds. To handle cases where the PCC approximation no longer holds, they adopted a data-driven approach with Gaussian process regressors to estimate forces. While the authors show robust detection of external disturbances, even in non steady-state conditions, they stop short of providing quantitative force estimates and do not show applications, such as a teach-mode. 

%% file: content/05-modeling.tex
\label{sec:ModelingTechniques}
In this section we describe the numerical model that is implemented in our simulation framework SOFA.

\subsection{Online Finite Element Modeling (FEM)}
We use the FEM, which yields the internal elastic forces $\mathbb{F}(\bm{q})$, given that the nodes of the FEM mesh are at positions $\bm{q}$. In SOFA, we use a formulation that accounts for the geometric non-linearities of the deformation and the material is characterized by the Hooke's law (Young's modulus and Poisson's ratio).
%
%
During each step $i$ of the simulation, a linearization of the internal forces is computed:
\begin{equation}
\label{eq:linearization}
\mathbb{F}(\bm{q_{i}}) \approx \mathbb{F}(\bm{q_{i-1}}) + K(\bm{q_{i-1}})d\bm{q},
\end{equation}
where $d\bm{q = q_{i}-q_{i-1}}$ is the displacement of the nodes and $K=\frac{\partial \mathbb{F}(\bm{q_{i-1}})}{\partial \bm{q}}$ is the tangential stiffness matrix for the current node positions $\bm{q}$. To complete the picture, external forces are included:
\begin{equation}
\label{eq:equilibrium}
0 = -K(\bm{q_{i-1}})d\bm{q} + \mathbb{P} + \mathbb{F}(\bm{q_{i-1}}) + H^{T}\bm{\lambda}.
\end{equation}
$H^{T}\bm{\lambda}$ is a vector that gathers boundary forces, such as contacts or external controlled inputs. The size of $\bm{\lambda}$ is equal to the number of rows in $H$ and to the number of actuators (contact forces, cables, etc.). $\mathbb{P}$ represents known external forces, such as gravity. Then, (\ref{eq:equilibrium}) is solved under the assumption of static equilibrium, delivering a motion that is a succession of quasi static states. Please refer to~\cite{duriez2013control} and \cite{coevoet2017software} for a more in-depth discussion about the FEM formulation used here.

\subsection{External Forces Actuation}
\label{subsec:ExternalContacts}


In the \textit{soft robot} plugin of SOFA, it is possible to model external force \emph{actuators} that simply apply forces on some point $p$ on the model. In this work, this force actuation is used to represent the pushing that will create a disturbance on the arm. The actuation magnitude is given by the variable $\lambda^{f}$. In this case, the forces are applied on the points $e_{1}$ and $e_{2}$ on the arm (see  Fig.~\ref{fig:OrientationEffector}). The direction $\vec{v}$ of force means a distribution on the three translational degrees of freedom, which is gathered in a row matrix $H = [...,\ \vec{v}^{T}\ ,...]$, so that the expression $H^{T}\lambda^{f}$ represents the force actuation in~(\ref{eq:equilibrium}). It is worth noting that in general it is possible to distribute the forces to a number of nodes, which could represent general pushing at arbitrary locations or areas on the manipulator. However, in this work, we only look at point-like pushing at the rigid segments.

\subsection{Pressure Actuation}
\label{subsec:PressureActuation}
For simulating the barymetric pressure, a variable $\lambda^{p}$ can be introduced. An internal cavity is represented by a triangulated surface mesh. Since pressure is force per area, the distribution of $\lambda^{p}$ on the nodes is calculated by taking into account the orientation and area of the triangles adjacent to each node, thus constructing the matrix $H$ in similar fashion than before. For further details, please refer to~\cite{coevoet2017software}.

\subsection{Fiber Reinforcement}
\label{subsec:FiberReinforcement}
\begin{figure}
    \centering
    \includegraphics[width = 0.60\linewidth,viewport=0 0 1255 553]{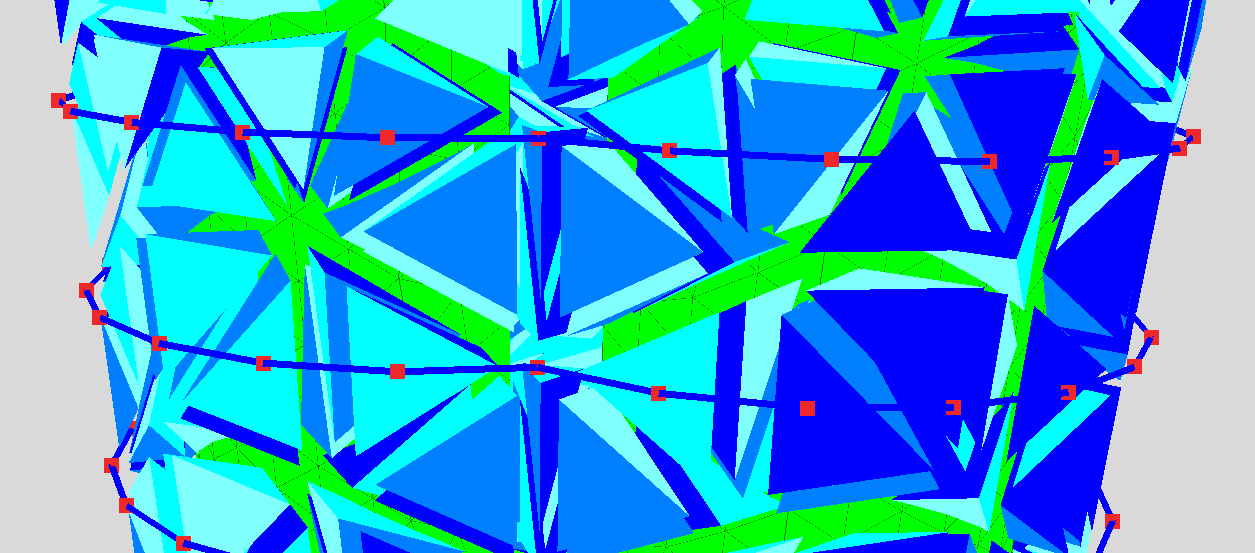}
    \caption{Fiber reinforcement modeled as a parallel sequence of segments  that are attached to the model. Each segment is a spring with a stiffness of \SI{1e4}{\newton\per\meter}. The total set of springs counteracts the radial expansion of the chamber during pressurization.}
    \label{fig:FiberReinforcement}
\end{figure}

It is well known that fiber reinforcement prevents large bulging of the pneumatic actuators, and helps in transferring the pressure effort more efficiently into bending or extension motion of soft manipulators. However, in the model of pressure actuation (see Sec.~\ref{subsec:PressureActuation}), it is not possible to account directly for the design of fiber-reinforced pressurized chambers. Therefore, we have modeled the effect of fiber reinforcement of the chambers by introducing stiff springs as shown in Fig~\ref{fig:FiberReinforcement}. Throughout the height of the chambers, sequences of segments are established that go around the circumference the arm. Each segment represents a spring, whose stiffness of stiffness \SI{1e4}{\newton\per\meter} is mapped to $K$ in (\ref{eq:equilibrium}) through a \emph{barycentric mapping}~\cite{coevoet2017software}.

\subsection{Orientation-Based State Observer}
\label{subsec:StateObserver}

\begin{figure}
    \centering
    \fontsize{8}{10}\selectfont
    \def\svgwidth{0.75\linewidth}
    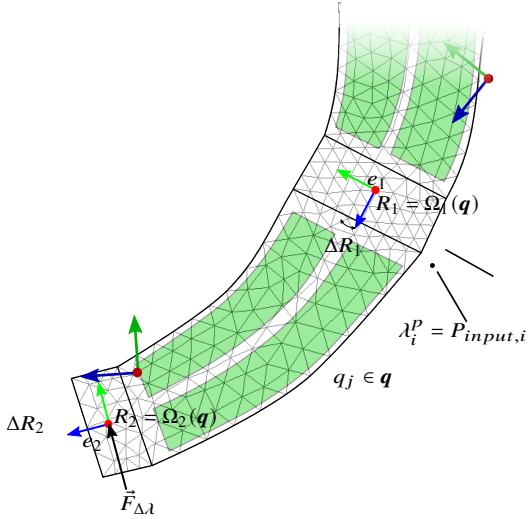
    \caption{A change in force $F_{\Delta \lambda}$ at $e_{1}$ induces a change in orientation $\Delta R_{1}$ and $\Delta R_{2}$ at both $e_{1}$ and $e_{2}$, according to (\ref{eq:DeltaR}). At the same time, the chambers are subject to pressure forces $\bm{\lambda}^{p}=\bm{P}_{input}$. Note that $F_{\Delta \lambda}$ also induces a translational change on $e_{1}$ and $e_{2}$. However, the translation of $e_{1}$ and $e_{2}$ is not a part of the observed state and is not taken into account by the effectors. The fiber reinforcements are not shown in this figure to avoid overcluttering.}
    \label{fig:OrientationEffector}
\end{figure}

In order to formulate the estimation of disturbance, we first need a description of the state of the arm. In SOFA, the state of the arm is completely described by the position of the FEM-nodes $\bm{q}$. However, it is prohibitive to equip the arm with as many sensors as FEM-nodes to observe its state. Thus, the state of the arm, as it will be observed by sensors, is described as the orientation at the two points $e_{1}$ and $e_{2}$ at the  rigid sections at the end of each section of fiber-reinforced chambers (see~Fig.~\ref{fig:OrientationEffector}). In our current setup, the arm is equipped with fiduciary markers, delivering the orientation at these points. In the future, we will equip each rigid section with IMUs that will provide an proprioceptive, i.\,e.\ embedded approach to sensing the orientations. In SOFA, the model is coupled to the real world through components called \emph{effectors}. This is a concept generalizing the well-known concept of \emph{end-effector} in robotics. In the optimization method used (see Sec.~\ref{subsec:ForceOptimization}), the actuation efforts that will minimize the difference between the state described by the effector component and a \emph{goal} value are calculated. In this case, the \emph{goal} are sensed orientation values. The realization of the \emph{orientation-effector} requires the implementation of a function
\begin{equation}
\label{eq:Omega}
R_{i}=\Omega_{i}(\bm{q}),
\end{equation}
mapping the current FEM-node positions $\bm{q}$ to the orientation of each rigid section $R_{i}$ at $e_{1}$ and $e_{2}$, and the partial derivative, relating the FEM-node movement to the change in the orientation, i.\,e.\ $\frac{\partial\Omega_{i}(\bm{q})}{\partial \bm{q}}$. Fig.~\ref{fig:OrientationEffector} illustrates the orientation-effector. 

\subsection{Finding Pressure and Disturbance Forces Through Optimization}
\label{subsec:ForceOptimization}

Using inverse problem solving \cite{coevoet2017software}, 
we can optimize for the  generalized actuation actuation forces that will minimize the distance between the effector values and the sensed values. In other words, we search for the actuation efforts that will best explain the observed orientations at each rigid segment. In this work, we consider two types of forces: actuation by pressurized air inside each one of the 6 chambers, $\Delta \bm{\lambda}^{p}= (\Delta\lambda^{p}_{1},\dots,\Delta\lambda^{p}_{6})^T$, and by pushing on the manipulator, $\Delta \bm{\lambda}^{f}= (\Delta\lambda^{f}_{1},\dots,\Delta\lambda^{f}_{4})^T$. The pushing forces are collinear with the local $x$ and $y$-axes at the points $e_{1}$ and $e_{2}$. At this time, there's no force to represent pulling/pushing, or twisting in the forward $z$-axis of each segment. A 2-axis flex sensor, like the one from BendLabs, does not allow observation of the necessary degree of freedom (stretching). However, an IMU will provide observability of additional degree of freedom, so we plan to include this in the future. Overall, all the actuation efforts are:
\begin{equation}
\label{eq:ActuationVariables}
\Delta \lambda=(\Delta \bm{\lambda}^{p}, \Delta \bm{\lambda}^{f})^T
\end{equation}. 
In this work, the first ones are assumed to be known. However, as we will see later, we solve the inverse problem of finding $\Delta \bm{\lambda}^{p}$ for given orientations in order to find the calibration between real pressures and simulated pressures (see Sec.~\ref{subsec:Calibration}). The estimation of the pushing forces $\Delta\bm{ \lambda^{f}}$ is the estimation of the disturbances.

To find the relation between the change in applied forces and the change in the orientation in simulation $\Delta R_{i,sim}$ of one orientation-effector we can write:
\begin{equation}
\label{eq:DeltaR}
\Delta R_{i,sim} = \frac{\partial\Omega_{i}(\bm{q})}{\partial \bm{q}} K^{-1} H^{T} \Delta \bm{\lambda}.
\end{equation}
Therefore, first the changes of forces are mapped to the corresponding nodes through $H^{T}$, as discussed in Sections~\ref{subsec:ExternalContacts} and ~\ref{subsec:PressureActuation}.
The tangential compliance matrix $K^{-1}$ transforms these forces to (FEM-)node-displacements, which finally can be mapped to changes in orientation through the derivative of $\Omega$ with respect to $\bm{q}$. For conciseness, we rewrite $\frac{\partial\Omega_{i}(\bm{q})}{\partial \bm{q}} K^{-1} H^{T}$ as $W_{i,ra}$, which is the matrix that directly maps changes in actuation force to changes in orientation. Now, we can formulate the optimization problem. We want to minimize the difference between the simulated orientations $\Delta R_{sim}=(W_{1,ra}\Delta \bm{\lambda}, \dots, W_{N,ra}\Delta \bm{\lambda} )^T=W_{ra} \Delta \bm{\lambda}$ and the real changes in orientation $\Delta R_{real} = (\Delta R_{1}, \dots, \Delta R_{N})^T$:
\begin{equation}
\label{eq:Optimization}
\Delta\bm{\lambda}^{*} = \argmin_{\Delta \bm{\lambda}}{\lVert W_{ra} \Delta \bm{\lambda} - \Delta R_{real} \rVert ^2}, 
\end{equation}
subject to some conditions, such as
\begin{equation}
    \Delta \bm{\lambda}_{max}\geq \Delta \bm{\lambda}\geq 0.
\end{equation}
If some of the efforts are known, this can be expressed as equality conditions. In particular, outside the calibrations scenario, actuation by pressurized air is known and thus the corresponding condition is
\begin{equation}
\label{eq:EqualityConstraints}
\Delta \bm{ \lambda}^{p} = \Delta \bm{P}_{input}.
\end{equation}

This quadratic program (QP) is solved using a QP-solver with the method presented in~\cite{coevoet2017software}.  
We obtain a value for all $\Delta \bm{\lambda}$ and in particular for the forces due to pushing.

%% file: images/Effector.pdf_tex
\begingroup%
  \makeatletter%
  \providecommand\color[2][]{%
    \errmessage{(Inkscape) Color is used for the text in Inkscape, but the package 'color.sty' is not loaded}%
    \renewcommand\color[2][]{}%
  }%
  \providecommand\transparent[1]{%
    \errmessage{(Inkscape) Transparency is used (non-zero) for the text in Inkscape, but the package 'transparent.sty' is not loaded}%
    \renewcommand\transparent[1]{}%
  }%
  \providecommand\rotatebox[2]{#2}%
  \newcommand*\fsize{\dimexpr\f@size pt\relax}%
  \newcommand*\lineheight[1]{\fontsize{\fsize}{#1\fsize}\selectfont}%
  \ifx\svgwidth\undefined%
    \setlength{\unitlength}{285.89240721bp}%
    \ifx\svgscale\undefined%
      \relax%
    \else%
      \setlength{\unitlength}{\unitlength * \real{\svgscale}}%
    \fi%
  \else%
    \setlength{\unitlength}{\svgwidth}%
  \fi%
  \global\let\svgwidth\undefined%
  \global\let\svgscale\undefined%
  \makeatother%
  \begin{picture}(1,1.06411962)%
    \lineheight{1}%
    \setlength\tabcolsep{0pt}%
    \put(0,0){\includegraphics[width=\unitlength,page=1,viewport=0 0 286 305]{Effector.pdf}}%
    \put(0.21697428,0.18397946){\color[rgb]{0,0,0}\makebox(0,0)[lt]{\lineheight{1.25}\smash{\begin{tabular}[t]{l}$R_{2}=\Omega_{2}(\bm{q})$\end{tabular}}}}%
    \put(0.75500292,0.62134175){\color[rgb]{0,0,0}\makebox(0,0)[lt]{\lineheight{1.25}\smash{\begin{tabular}[t]{l}$R_{1}=\Omega_{1}(\bm{q})$\end{tabular}}}}%
    \put(0.66860969,0.26713973){\color[rgb]{0,0,0}\makebox(0,0)[lt]{\lineheight{1.25}\smash{\begin{tabular}[t]{l}$q_{j}\in \bm{q}$\end{tabular}}}}%
    \put(0,0){\includegraphics[width=\unitlength,page=2,viewport=0 0 212 206]{Effector.pdf}}%
    \put(0.23106524,0.00526099){\color[rgb]{0,0,0}\makebox(0,0)[lt]{\lineheight{1.25}\smash{\begin{tabular}[t]{l}$\vec{F}_{\Delta \lambda}$\end{tabular}}}}%
    \put(0,0){\includegraphics[width=\unitlength,page=3,viewport=0 0 34 60]{Effector.pdf}}%
    \put(-0.00222784,0.17037195){\color[rgb]{0,0,0}\makebox(0,0)[lt]{\lineheight{1.25}\smash{\begin{tabular}[t]{l}$\Delta R_{2}$\end{tabular}}}}%
    \put(0.65058369,0.53496788){\color[rgb]{0,0,0}\makebox(0,0)[lt]{\lineheight{1.25}\smash{\begin{tabular}[t]{l}$\Delta R_{1}$\end{tabular}}}}%
    \put(0,0){\includegraphics[width=\unitlength,page=4,viewport=0 0 282 301]{Effector.pdf}}%
    \put(0.15337486,0.13422796){\color[rgb]{0,0,0}\makebox(0,0)[lt]{\lineheight{1.25}\smash{\begin{tabular}[t]{l}$e_{2}$\end{tabular}}}}%
    \put(0.7380104,0.67295459){\color[rgb]{0,0,0}\makebox(0,0)[lt]{\lineheight{1.25}\smash{\begin{tabular}[t]{l}$e_{1}$\end{tabular}}}}%
    \put(0.8022873,0.36045798){\color[rgb]{0,0,0}\makebox(0,0)[lt]{\lineheight{1.25}\smash{\begin{tabular}[t]{l}$\lambda^{p}_{i}=P_{input,i}$\end{tabular}}}}%
    \put(0,0){\includegraphics[width=\unitlength,page=5,viewport=0 0 228 124]{Effector.pdf}}%
  \end{picture}%
\endgroup%

%% file: content/06-experiments.tex
\subsection{Setup}

Our setup comprises of a 2-segment soft robotic arm with fiber-reinforced pneumatic actuators as shown in Fig.~ \ref{fig:overall-system}.  Each segment of the arm is made up of three individually fiber-reinforced silicone elastomer air-chambers (Shore hardness: 10A) glued and held together by an enveloping layer of the same material. The arm is designed tapered and the top and bottom soft segments with air cavities are sandwiched between relatively rigid segments made of a different type of silicone elastomer (Shore hardness: 30A). This rigid intermediate segment houses the pneumatic pathways between the different segments. To track the pose of robotic arm, we use reflective markers attached to the rigid sections at the top, middle and at the bottom that are tracked by a $Qualisys^{TM}$ motion-capture system.

The set of markers attached to each segment are grouped together in the tracking system into a rigid frames: base (B), intermediate (I), and tip (T). The motion capture system outputs the measured (m) pose transformations of these frame relative to an arbitrary world frame (W) as $X_{(m)WB}$, $X_{(m)WI}$, and $X_{(m)WT}$. From these, we can compute the relative transformations that represent the the orientation in the arm segments:
$ X_{(m)BI} = X^{-1}_{(m)WB} \times X_{(m)WI} $
and 
$ X_{(m)IT} = X^{-1}_{(m)WI} \times X_{(m)WT} $.

From the simulation running on SOFA, we obtain the nominal (n) pose transformations for a given set of chamber pressure values $X_{(n)BI}$, $X_{(n)IT}$.

We then compare the nominal and measured pose values with zero actuation (pressure = $\SI{0}{\pascal}$) to compute a rectification pose transform ($\Delta$) for each segment that corrects for small errors in measurement due to the fabrication process:
$ X_{(\Delta)BI} = X^{-1}_{(n)BI} \times X_{(m)BI} $
and 
$ X_{(\Delta)IT} = X^{-1}_{(n)IT} \times X_{(m)IT} $. We apply this transformation to correct for systemic errors in pose measurement and compute rectified pose transformation (r) corresponding to each arm segment:
$X_{(r)BI} = X_{(\Delta)BI} \times X_{(m)BI}$ and
$X_{(r)IT} = X_{(\Delta)IT} \times X_{(m)IT}$.

For external force measurements, we used a 1-axis force gauge with range \SI{5}{\kilo\gram} and  \SI{1}{\gram} resolution. The force sensor was rigidly mounted to a frame such that the axis of measurement aligns with the motion of the point on the arm at which force measurement is being performed.  The sensor it self is the attached to the arm using a thin piece of cord.

\subsection{Calibration using Inverse Dynamics}
\label{subsec:Calibration}

The mesh used to represent SoPrA for the FEM simulation has 1201 nodes and 3869 tetrahedral elements. While these parameters allow an execution rate of $\approx\SI{12}{\hertz}$, they are far away from FEM-convergence. Thus, for optimal behavior, we can not expect to set the material properties from a datasheet or test-bench results, but need to do so empirically, by calibration. The parameter subject to tuning is Young's Modulus $E$ (Poisson's Ratio is fixed at $0.45$).

We found that there are two types of calibration that are needed, depending on whether we aim for 1) correct force input (air pressure) and correct orientation output or 2) correct force output and correct orientation output.

The first calibration will give us a one-to-one relationship between the pressure values coming from the electronic valve banks controlling SoPrA and the pressure value in simulation. Using the optimization presented in Sec.~\ref{subsec:ForceOptimization}, we solve for the pressure values in simulation that bring the arm as close as possible to the orientation determined at the two rigid segments by the sensors (optical tracking). Concretely, a series of experiments is performed where the pressures in each of the six chambers were increased gradually from \SI{0}{\pascal} to \SI{65}{\kilo\pascal} (which was 
determined to be the maximum safe pressure without permanently damaging the actuator), and recorded the real and simulated pressures corresponding to each chamber. Comparing the real and simulated pressures yields a factor $\alpha$ that is used to scale the Young's modulus $E$ and a vector scaling factor $\bm{\nu}  = [\nu_1, \nu_2, \dots, \nu_{6}]$ that accounts for discrepancies that arise due to the very manual nature of the fabrication process.
\begin{equation}
    \label{eq:PressureCalibrationFactors}
    \frac{1}{6} \sum_{i=1}^{6}{\nu_i} \approx 1
\end{equation}

Since SoPrA is redundant, solving for the pressures to achieve a given pose of the arm, specified in terms of orientation angles of the two segments has infinitely many solutions. To constrain the problem, during this calibration process, we limit the solution space to the two chambers, one in each segment, that are being actuated. Otherwise, the inverse-dynamics-based optimization might find solutions that have a different actuation pattern than the real actuation, thus potentially distorting the data for calibration.

In the second calibration, the same procedure is followed, only that now Young's Modulus is adjusted so that the model output in terms of force is consistent with the experiments. After this, the factors in (\ref{eq:PressureCalibrationFactors}) will no longer have an average of one. The model is still consistent in terms of estimation of orientations, but the input pressures now need to be adjusted by an empirical factor.

\subsection{Validation of Inverse Dynamics-based Calibration}
\label{subsec:Ik-validation}
To validate the parameters obtained above, we perform a leave-one-out random actuation experiment where we pick two out of three chambers per segment and and apply randomly sampled pressure values to these chambers. This leave-one-out procedure is required since the model, in this case, only receives orientation measurements and cannot distinguish between two cases where the pose of the arm is the same with different stiffness values. The SOFA model then is used to compute estimated pose of the arm from measurements. We quantify performance and validate the calibration procedure by computing position and orientation error between measured and estimated poses as shown in Table \ref{tab:ik_validation-results}.

\subsection{Teach mode}
\label{subsec:TeachMode}
We apply the Young's modulus and pressure scaling factors computed in the above calibration process to implement a "teach mode" feature on the arm. We set up the simulation to estimate pressures required to reach a given arm pose specified by orientation angles. The user moves the robot manually to a desired pose $X_{(d)BI}$ and while the arm is being moved, estimated pressures are computed and updated in parallel. Once commanded by the user, teach mode ramps up pressures in the arm to those estimated by the simulation to maintain this user desired pose. It is important to note that unlike the calibration scenario, we do not have knowledge about which chambers should be actuated, because only orientation information is provided. Thus, the solver can decide to use up to six chambers to reach the target orientations, potentially reducing the pose error due to higher actuation ability. 

The experiment was repeated with various and the results are shown in \ref{tab:teach_mode-results}. Position error is computed as the distance between the measured final pose reached by the arm compared to the user's desired pose and orientation error is the shortest angular distance between the two poses.

\begin{figure}[!t]
    \centering
    \includegraphics[width = 0.75\linewidth,viewport=0 0 1822 1220]{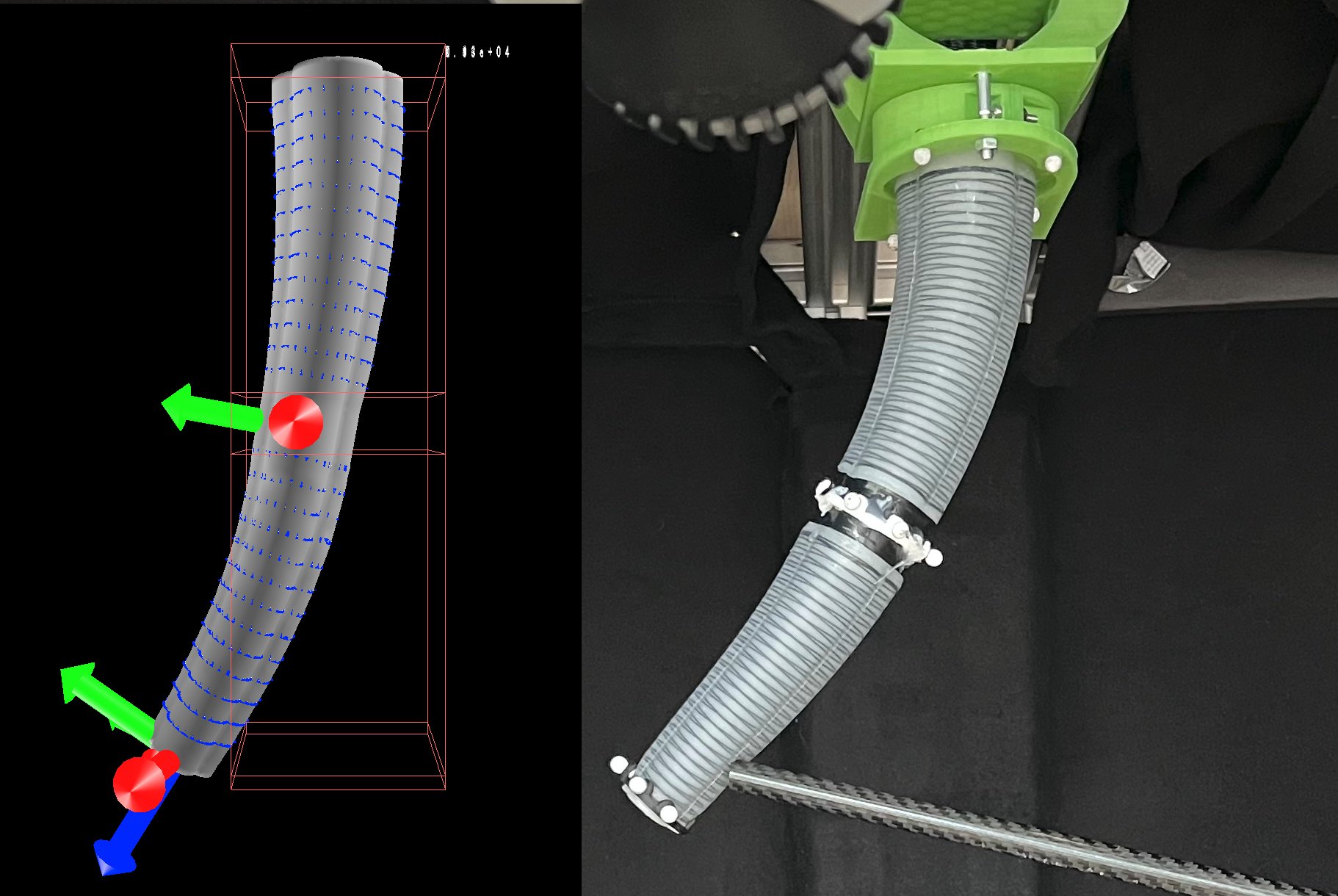}
    \caption{Teach mode: User moves the arm to desired pose. Model is used to estimate pressures required to reach desired and pose and these estimated pressures are then scaled and sent to the arm as commanded pressures}
    \label{fig:exp-teach_mode}
\end{figure}

\subsection{Disturbance estimation - chambers not actuated}
\label{subsec:DisturbanceEstNotActuated}
As described in Sec.~\ref{subsec:Calibration}, the calibration parameters are found in order for the model to deliver properly scaled force outputs ( calibration 2)). A SOFA scene is configured to estimate disturbance in at tip of the arm. The simulation takes as inputs the measured orientation and measured pressures, which in this case are all equal to \SI{0}{\pascal}. The SOFA model computes estimated poses of the two frames attached to the intermediate rigid section and the tip of the arm and these are compared with the measured orientation from motion capture. The disturbance magnitude returned minimizes the difference measured and estimated orientation. Quantitative results form this experiment are presented in Fig.\ref{fig:force-est-unactuated-arm} and in Table \ref{tab:force_est-results}.

\subsection{Disturbance estimation - actuated arm}
\label{subsec:DisturbanceEstActuated}
\begin{figure}[!t]
    \centering
    \includegraphics[width = 0.75\linewidth,viewport=0 0 2206 1732]{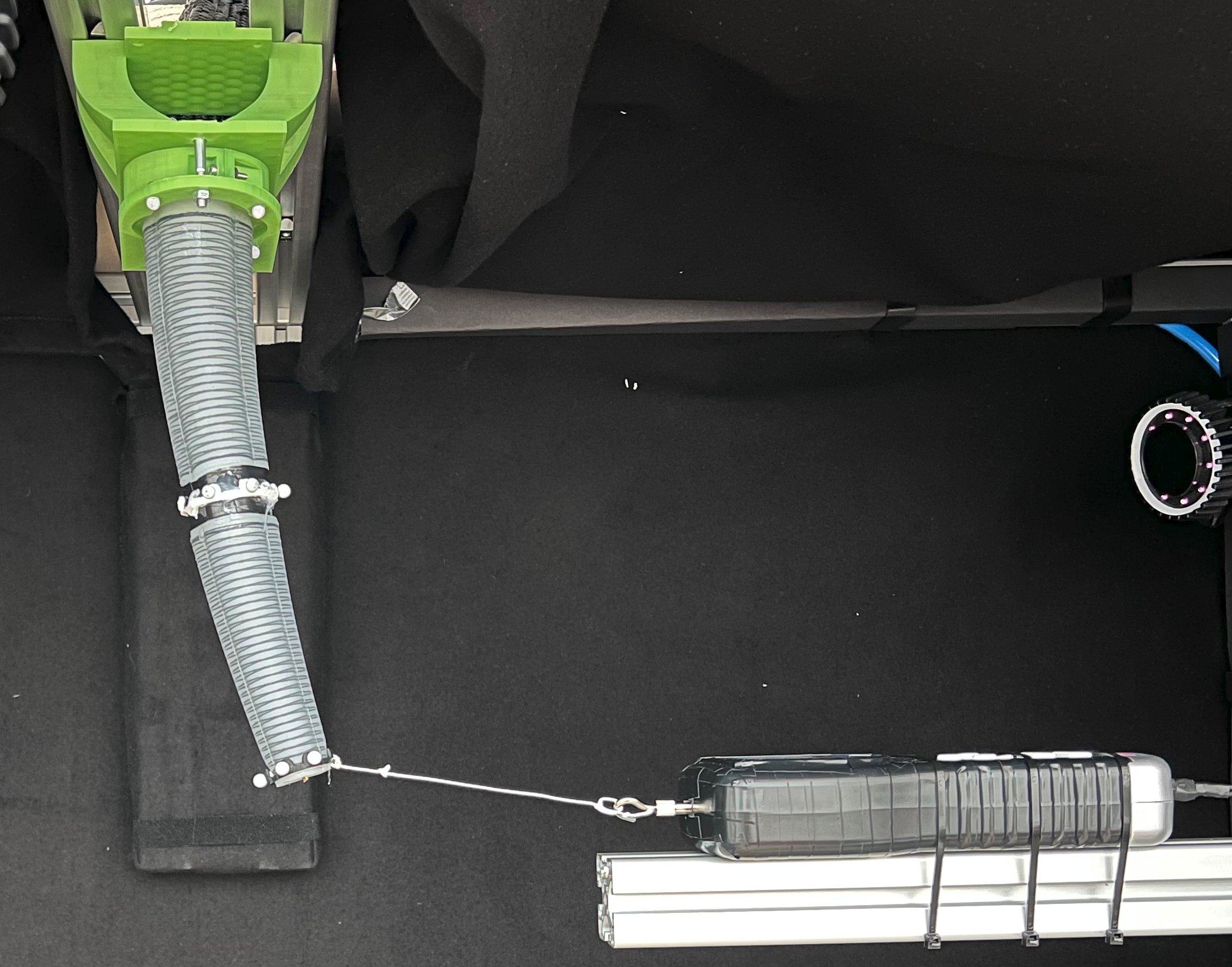}
    \caption{Disturbance estimation with the arm not pressurized. The sensor is moved away from the arm in small steps and the force measured by the sensor, arm segment orientation, and measure pressures are logged}
    \label{fig:experiment-force_est}
\end{figure}

The disturbance estimation experiment is repeated with the arm pressurized. The experiment is setup as shown in Fig. \ref{fig:experiment-force_est}. The tip of the arm is attached to the force gauge and pressure in the chambers is ramped up gradually in steps such that the tip of the arm tries to move away from the force gauge. The motion is, however, limited by the cord attaching it to the force gauge and the force exerted on the sensor is measured. Measured pressure values are applied in the arm chambers using equality constraints of (\ref{eq:EqualityConstraints}), and the SOFA simulation is configured to estimate the external disturbances acting on the arm that would explain the discrepancy between measured and estimated orientation angles. Results from the experiment are shown in Fig. \ref{fig:force-est-actuated-arm} and in Table \ref{tab:force_est-results}. 

%% file: content/07-results.tex
\begin{table}[]
\centering
\caption{Results from validation of the inverse dynamics-based calibration.\label{tab:valid-results}}
\begin{tabular}{@{}lcc@{}}
\toprule
Error statistics  & \thead{Position error (m)} & \thead{Orientation error (deg)} \\ 
\midrule
min    & 0.0061             & 2.0930              \\ 
max    & 0.0593             & 26.5407             \\ 
mean   & 0.0365             & 13.7210             \\ 
std    & 0.0163             & 6.7822              \\ 
\bottomrule
\end{tabular}

\label{tab:ik_validation-results}
\end{table}

\begin{table}[]
\centering
\caption{Teach mode results (from 15 experiments)\label{tab:teach_mode-results}}
\begin{tabular}{lcc}
\toprule
Error statistics  & \thead{Position error (m)} & \thead{Orientation error (deg)} \\
\midrule
min    & 0.0017             & 1.33                 \\ 
max    & 0.0135             & 11.36                 \\ 
mean   & 0.0052             & 4.04                 \\ 
std    & 0.0037             & 3.01                 \\ 
\bottomrule
\end{tabular}
\end{table}

\begin{table}[]
\centering
\caption{Force estimation results}
\begin{tabular}{@{}lcc@{}}
\toprule
Force estimation error (N)  & \thead{Not Pressurized} & \thead{Pressurized} \\
\midrule
min                    & 7.585e-07                & 3.349e-06            \\
max                    & 0.059                    & 0.177                \\
mean                   & 0.007                    & 0.0238               \\
std                    & 0.008                    & 0.022                \\
experiment range       & 0.66708                  & 1.98162              \\
\bottomrule
\end{tabular}
\label{tab:force_est-results}
\end{table}

\begin{figure}[!t]
    \centering
    \includegraphics[width=\columnwidth,viewport=0 0 233 233]{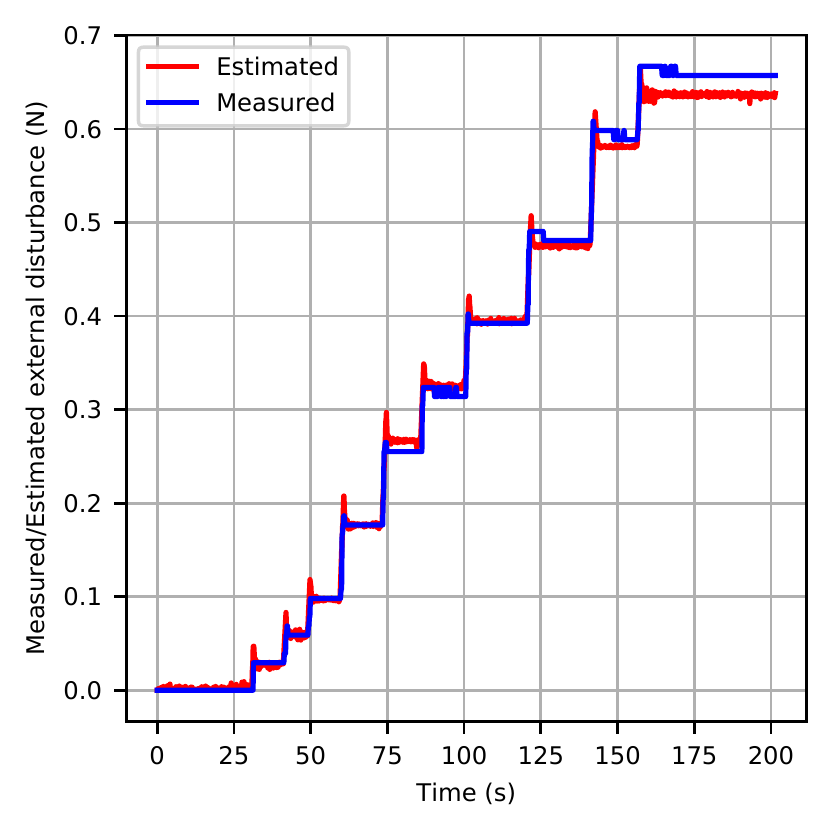}
    \caption{Measured and estimated disturbance over time (Not Pressurized)}
    \label{fig:force-est-unactuated-arm}
\end{figure}

\begin{figure}[!t]
    \centering
    \includegraphics[width=\columnwidth,viewport=0 0 232 233]{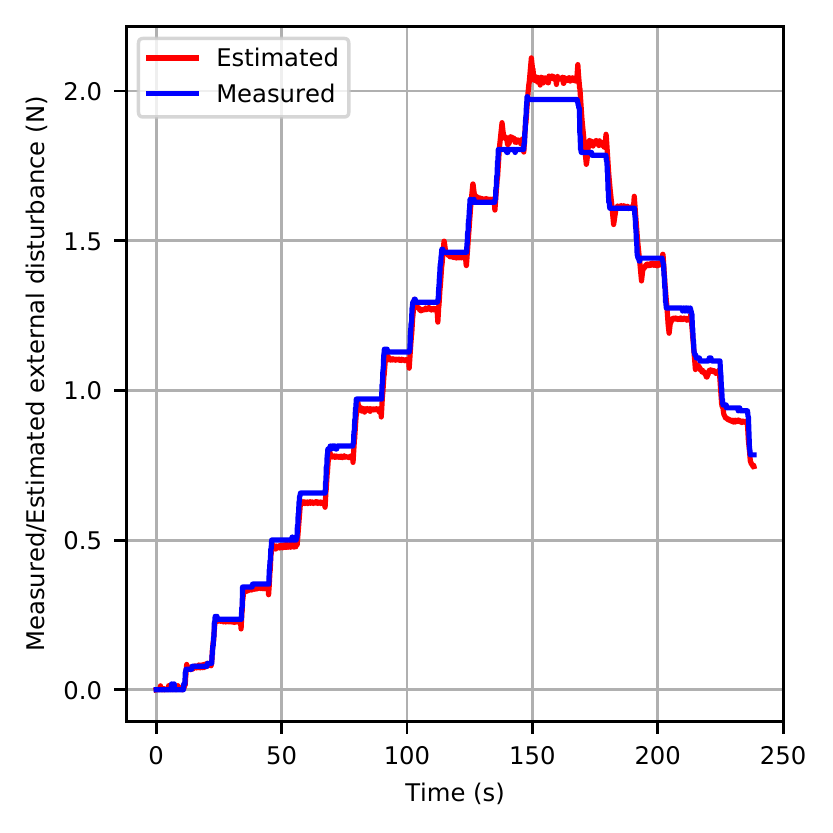}
    \caption{Measured and estimated disturbance over time (Pressurized)}
    \label{fig:force-est-actuated-arm}
\end{figure}

%% file: content/09-conclusion.tex
In this paper, we have presented a model-based approach for disturbance estimation for soft robots. Using the FEM, we have modeled a soft manipulator arm with fiber-reinforced pressurized chambers as actuators and orientation sensing as means of proprioception. The proposed modeling approach allows the seamless integration of actuation efforts with orientation sensing to yield disturbance estimation. We have performed several experiments to validate our approach. First, it is shown that the robust state modeling allows solving the task of inverse dynamics, i.\,e.\ finding the pressures needed to let the robot reach desired orientations of its intermediate segments. This can be used to realize the so called ``teach mode'', that is, let the user program the robot by moving the robot arm to the desired pose. Second, force estimation experiments are carried out, without additional actuation and with actuation. The results in both experiments are quite satisfactory, as the robot is able to reach the target poses with good accuracy and the estimated forces are correct. That being said, different calibrations are needed to compensate for different scaling factors when the target is expressed in terms of orientation or forces.

In terms of future work, we are optimistic about being able to further generalize the proposed approach. One of the first priorities is to integrate IMUs into the segments of the arm to not depend on external sensing anymore. Furthermore, equipping the arm with a sensor skin will allow the automatic detection of location of disturbances. Therefore, the arm's potential for deployment in human-centered environment would increase further. The augmented sensorial capabilities will also play an important role in contact-intense exploration tasks. 

%% file: Bibliography.tex

%% file: root.bbl
\begin{thebibliography}{10}
\providecommand{\url}[1]{#1}
\csname url@samestyle\endcsname
\providecommand{\newblock}{\relax}
\providecommand{\bibinfo}[2]{#2}
\providecommand{\BIBentrySTDinterwordspacing}{\spaceskip=0pt\relax}
\providecommand{\BIBentryALTinterwordstretchfactor}{4}
\providecommand{\BIBentryALTinterwordspacing}{\spaceskip=\fontdimen2\font plus
\BIBentryALTinterwordstretchfactor\fontdimen3\font minus
  \fontdimen4\font\relax}
\providecommand{\BIBforeignlanguage}[2]{{%
\expandafter\ifx\csname l@#1\endcsname\relax
\typeout{** WARNING: IEEEtran.bst: No hyphenation pattern has been}%
\typeout{** loaded for the language `#1'. Using the pattern for}%
\typeout{** the default language instead.}%
\else
\language=\csname l@#1\endcsname
\fi
#2}}
\providecommand{\BIBdecl}{\relax}
\BIBdecl

\bibitem{rus_design_2015}
\BIBentryALTinterwordspacing
D.~Rus and M.~T. Tolley, ``\BIBforeignlanguage{english}{Design, fabrication and
  control of soft robots},'' \emph{\BIBforeignlanguage{english}{Nature}}, vol. 521,
  no. 7553, pp. 467--475, May 2015, number: 7553 Publisher: Nature Publishing
  Group. [Online]. Available: \url{https://www.nature.com/articles/nature14543}
\BIBentrySTDinterwordspacing

\bibitem{hawkes_hard_2021}
\BIBentryALTinterwordspacing
E.~W. Hawkes, C.~Majidi, and M.~T. Tolley, ``\BIBforeignlanguage{english}{Hard
  questions for soft robotics},'' \emph{\BIBforeignlanguage{english}{Science
  Robotics}}, vol.~6, no.~53, Apr. 2021, publisher: Science Robotics Section:
  Viewpoint. [Online]. Available:
  \url{https://robotics.sciencemag.org/content/6/53/eabg6049}
\BIBentrySTDinterwordspacing

\bibitem{della_santina_dynamic_2018}
C.~Della~Santina, R.~K. Katzschmann, A.~Biechi, and D.~Rus, ``Dynamic control
  of soft robots interacting with the environment,'' in \emph{2018 {IEEE}
  {International} {Conference} on {Soft} {Robotics} ({RoboSoft})}, Apr. 2018,
  pp. 46--53.

\bibitem{homberg_haptic_2015}
B.~S. Homberg, R.~K. Katzschmann, M.~R. Dogar, and D.~Rus, ``Haptic
  identification of objects using a modular soft robotic gripper,'' in
  \emph{2015 {IEEE}/{RSJ} {International} {Conference} on {Intelligent}
  {Robots} and {Systems} ({IROS})}, Sep. 2015, pp. 1698--1705.

\bibitem{coad_vine_2020}
M.~M. Coad, L.~H. Blumenschein, S.~Cutler, J.~A.~R. Zepeda, N.~Naclerio,
  H.~El-Hussieny, U.~Mehmood, J.~Ryu, E.~Hawkes, and A.~Okamura, ``Vine
  {Robots}: {Design}, {Teleoperation}, and {Deployment} for {Navigation} and
  {Exploration},'' \emph{IEEE Robotics \& Automation Magazine}, 2020.

\bibitem{n_kuppuswamy_soft-bubble_2021}
{N. Kuppuswamy}, {A. Alspach}, {A. Uttamchandani}, {S. Creasey}, {T. Ikeda},
  and {R. Tedrake}, ``Soft-bubble grippers for robust and perceptive
  manipulation,'' in \emph{2020 {IEEE}/{RSJ} {International} {Conference} on
  {Intelligent} {Robots} and {Systems} ({IROS})}, Jan. 2021, pp. 9917--9924,
  journal Abbreviation: 2020 IEEE/RSJ International Conference on Intelligent
  Robots and Systems (IROS).

\bibitem{lessing_soft_2018}
\BIBentryALTinterwordspacing
J.~A. Lessing, R.~R. Knopf, C.~E. Vause, and K.~Alcedo,
  ``\BIBforeignlanguage{english}{Soft robotic actuator attachment hub and grasper
  assembly, reinforced actuators, and electroadhesive actuators},'' US Patent
  US10\,118\,301B2, Nov., 2018. [Online]. Available:
  \url{https://patents.google.com/patent/US10118301B2/en}
\BIBentrySTDinterwordspacing

\bibitem{graetzel_robotic_2019}
C.~F. Graetzel, A.~Sheehy, and D.~P. Noonan, ``Robotic bronchoscopy drive mode
  of the {Auris} {Monarch} platform*,'' in \emph{2019 {International}
  {Conference} on {Robotics} and {Automation} ({ICRA})}, May 2019, pp.
  3895--3901, iSSN: 2577-087X.

\bibitem{rogers_methods_2018}
\BIBentryALTinterwordspacing
T.~W. Rogers and D.~Q. Larkin, ``\BIBforeignlanguage{english}{Methods and apparatus
  to shape flexible entry guides for minimally invasive surgery},'' US Patent
  US9\,962\,066B2, May, 2018. [Online]. Available:
  \url{https://patents.google.com/patent/US9962066B2/en}
\BIBentrySTDinterwordspacing

\bibitem{nguyen_fabric_2019}
\BIBentryALTinterwordspacing
P.~H. Nguyen, I.~I.~B. Mohd, C.~Sparks, F.~L. Arellano, W.~Zhang, and
  P.~Polygerinos, ``Fabric {Soft} {Poly}-{Limbs} for {Physical} {Assistance} of
  {Daily} {Living} {Tasks},'' \emph{arXiv:1903.07852 [cs]}, Mar. 2019, arXiv:
  1903.07852. [Online]. Available: \url{http://arxiv.org/abs/1903.07852}
\BIBentrySTDinterwordspacing

\bibitem{toshimitsu2021sopra}
Y.~Toshimitsu, K.~W. Wong, T.~Buchner, and R.~Katzschmann, ``Sopra: Fabrication
  \& dynamical modeling of a scalable soft continuum robotic arm with
  integrated proprioceptive sensing,'' in \emph{2021 IEEE/RSJ International
  Conference on Intelligent Robots and Systems (IROS)}.\hskip 1em plus 0.5em
  minus 0.4em\relax IEEE, 2021, pp. 653--660.

\bibitem{navarro2020model}
S.~{Escaida Navarro}, S.~Nagels, H.~Alagi, L.-M. Faller, O.~Goury,
  T.~Morales-Bieze, H.~Zangl, B.~Hein, R.~Ramakers, W.~Deferme, G.~Zheng, and
  C.~Duriez, ``A model-based sensor fusion approach for force and shape
  estimation in soft robotics,'' \emph{IEEE Robotics and Automation Letters},
  vol.~5, no.~4, pp. 5621--5628, 2020.

\bibitem{navarro2021bio}
S.~E. Navarro, S.~S. Dhaliwal, M.~S. Lopez, S.~Wilby, A.~L. Palmer, W.~Polak,
  R.~Merzouki, and C.~Duriez, ``A bio-inspired active prostate phantom for
  adaptive interventions,'' \emph{IEEE Transactions on Medical Robotics and
  Bionics}, 2021.

\bibitem{coevoet2017software}
E.~Coevoet, T.~Morales-Bieze, F.~Largilliere, Z.~Zhang, M.~Thieffry,
  M.~Sanz-Lopez, B.~Carrez, D.~Marchal, O.~Goury, J.~Dequidt \emph{et~al.},
  ``Software toolkit for modeling, simulation, and control of soft robots,''
  \emph{Advanced Robotics}, vol.~31, no.~22, pp. 1208--1224, 2017.

\bibitem{polygerinos2015modeling}
P.~Polygerinos, Z.~Wang, J.~T. Overvelde, K.~C. Galloway, R.~J. Wood,
  K.~Bertoldi, and C.~J. Walsh, ``Modeling of soft fiber-reinforced bending
  actuators,'' \emph{IEEE Transactions on Robotics}, vol.~31, no.~3, pp.
  778--789, 2015.

\bibitem{tapia2020makesense}
J.~Tapia, E.~Knoop, M.~Mutn{\`y}, M.~A. Otaduy, and M.~B{\"a}cher, ``Makesense:
  Automated sensor design for proprioceptive soft robots,'' \emph{Soft
  robotics}, vol.~7, no.~3, pp. 332--345, 2020.

\bibitem{yirmibesoglu2016hybrid}
O.~D. Yirmibesoglu and Y.~Menguc, ``Hybrid soft sensor with embedded imus to
  measure motion,'' in \emph{2016 IEEE International Conference on Automation
  Science and Engineering (CASE)}.\hskip 1em plus 0.5em minus 0.4em\relax IEEE,
  2016, pp. 798--804.

\bibitem{hellebrekers2018liquid}
T.~Hellebrekers, K.~B. Ozutemiz, J.~Yin, and C.~Majidi, ``Liquid
  metal-microelectronics integration for a sensorized soft robot skin,'' in
  \emph{2018 IEEE/RSJ International Conference on Intelligent Robots and
  Systems (IROS)}.\hskip 1em plus 0.5em minus 0.4em\relax IEEE, 2018, pp.
  5924--5929.

\bibitem{hughes2020sensing}
J.~Hughes, F.~Stella, C.~D. Santina, and D.~Rus, ``Sensing soft robot shape
  using imus: An experimental investigation,'' in \emph{International Symposium
  on Experimental Robotics}.\hskip 1em plus 0.5em minus 0.4em\relax Springer,
  2020, pp. 543--552.

\bibitem{scimeca2019model}
L.~Scimeca, J.~Hughes, P.~Maiolino, and F.~Iida, ``Model-free soft-structure
  reconstruction for proprioception using tactile arrays,'' \emph{IEEE Robotics
  and Automation Letters}, vol.~4, no.~3, pp. 2479--2484, 2019.

\bibitem{della2020data}
C.~Della~Santina, R.~L. Truby, and D.~Rus, ``Data--driven disturbance observers
  for estimating external forces on soft robots,'' \emph{IEEE Robotics and
  automation letters}, vol.~5, no.~4, pp. 5717--5724, 2020.

\bibitem{duriez2013control}
C.~Duriez, ``Control of elastic soft robots based on real-time finite element
  method,'' in \emph{Robotics and Automation (ICRA), 2013 IEEE International
  Conference on Robotics and Automation}.\hskip 1em plus 0.5em minus
  0.4em\relax IEEE, 2013, pp. 3982--3987.

\end{thebibliography}
